\newcommand\CLASSINPUTinnersidemargin{0.75in}
\newcommand\CLASSINPUToutersidemargin{0.75in}
\newcommand\CLASSINPUTtoptextmargin{0.75in}
\newcommand\CLASSINPUTbottomtextmargin{0.75in}
\documentclass[conference,letterpaper]{IEEEtran}
\IEEEoverridecommandlockouts
\usepackage{cite}
\usepackage{amsmath,amssymb,amsfonts}
\usepackage{algorithm}
\usepackage{algorithmic}
\usepackage{graphicx}
\usepackage{textcomp}
\usepackage{booktabs}
\usepackage{multirow}
\usepackage{makecell}
\usepackage{placeins}
\usepackage[table]{xcolor} 
\usepackage[colorlinks=true,linkcolor=blue,citecolor=blue,urlcolor=blue]{hyperref}
\newcommand{\score}[2]{#1\textcolor[gray]{0.5}{\small{$\,\pm#2$}}}
\newcommand{\bestscore}[2]{\textbf{#1}\textcolor[gray]{0.5}{\small{$\,\pm#2$}}}
\newcommand{\scoremean}[1]{#1}
\def\BibTeX{{\rm B\kern-.05em{\sc i\kern-.025em b}\kern-.08em
    T\kern-.1667em\lower.7ex\hbox{E}\kern-.125emX}}
\begin{document}

\title{ARP: Enhancing Quantized Skill Abstractions via Visual Alignment and Iterative Refinement for Robotic Manipulation}

\makeatletter
\def\IEEEtitletopspaceextra{18pt}
\makeatother

\author{
Yuntian Wang$^{1,2}$, Zesheng Jia$^{1,2}$, Yuhui Duan$^{1,2}$, Qibing Wang$^{3}$,\\
Yang Liu$^{4}$, Song Wang$^{5}$, Siao Liu$^{1,2,*}$, and Jin Wang$^{1,2,*}$
\thanks{$^{1}$School of Future Science and Engineering, Soochow University;
$^{2}$Key Laboratory of General Artificial Intelligence and Large Models in Provincial Universities, Soochow University;
$^{3}$College of Mechanical and Electronic Engineering, China Jiliang University;
$^{4}$College of Electronics and Information Engineering, Tongji University;
$^{5}$Leju Robotics. $^{*}$Corresponding authors: wjin1985@suda.edu.cn; saliu@suda.edu.cn.\par
This work was supported in part by the National Natural Science Foundation of China (62072321), the Science and Technology Program of Jiangsu Province (BZ2024062), the Natural Science Foundation of the Jiangsu Higher Education Institutions of China (22KJA520007), Suzhou Planning Project of Science and Technology (2023ss03, SYG2025129, SNG2025010) and Key Laboratory of General Artificial Intelligence and Large Models in Provincial Universities, Soochow University.\par
Project page available at: \url{https://arp-policy.github.io/}.}
}

\maketitle

\begin{abstract}
Learning visuomotor policies for long-horizon manipulation remains a fundamental challenge. Recent skill-based imitation learning methods based on discrete quantization have shown promising results by representing complex behaviors as temporally extended skills. However, most existing approaches primarily encode action trajectories into latent skills, yielding weak visual-semantic grounding and limiting the ability to leverage visual observations for skill selection. Moreover, discrete tokenization inevitably incurs precision errors during continuous action generation. To alleviate these issues, we propose \textbf{Aligned Refinement Policy (ARP)}, a discrete-skill framework that couples semantic grounding with execution-level refinement. Specifically, ARP introduces (i) a \textbf{visual--action alignment} objective that contrastively aligns visual embeddings with pre-quantized action representations in a shared latent space while preserving a state-independent skill decoder, and (ii) a lightweight \textbf{Iterative Residual Head (IRH)} that performs a two-step refinement to recover fine-grained control for precise execution. Extensive experiments show that ARP achieves state-of-the-art performance on the LIBERO and Meta-World benchmarks. Moreover, real-robot experiments on the Kuavo 4 Pro humanoid platform further validate its effectiveness, yielding consistent performance gains over several baselines on two challenging manipulation tasks.
\end{abstract}


\section{Introduction}
Vision-based manipulation aims to learn decision policies that map high-dimensional observations to continuous robot actions. Recently, imitation learning (IL) methods have achieved substantial progress by directly training visuomotor policies from expert demonstrations~\cite{act,diffusionpolicy,liu2025grasp,jin2026umi}. In our work, we focus on a more challenging setting: \textit{long-horizon manipulation}, where tasks require the composition of multiple contact-rich primitives over extended horizons. In this regime, minor perceptual or control errors can temporally compound, leading to error accumulation and degraded performance~\cite{imitation_progress_taxonomies_challenges}.

One promising solution is to decompose long-horizon trajectories into reusable skills, providing structure for multi-task composition~\cite{ju2024rethinking}.
Existing skill learning methods can be roughly divided into two categories.
One line learns compressed, observation-based skill representations that are shared across tasks, deriving discrete skills from language paired with visual inputs~\cite{garg2022lisa,liang2024skilldiffuser}.
Although this provides semantic grounding, the use of observations can entangle the latent skills with appearance-specific shortcuts, blurring reusable action primitives with the scene context.
Another line focuses on extracting latent skills from raw action sequences using discrete representations such as VQ-VAE~\cite{vqvae} and discrete action tokenizers~\cite{vqbet,quest,discrete_policy}.
However, while these action-only tokenizers yield compact and reusable motion primitives, they are often detached from visual semantics, leading to semantic ambiguity under visual and language variation.
Moreover, discretization inevitably loses fine-grained geometric details, which can accumulate over long-horizon execution and degrade precision. 

\begin{figure}[!t]
  \centering
  \includegraphics[width=\columnwidth]{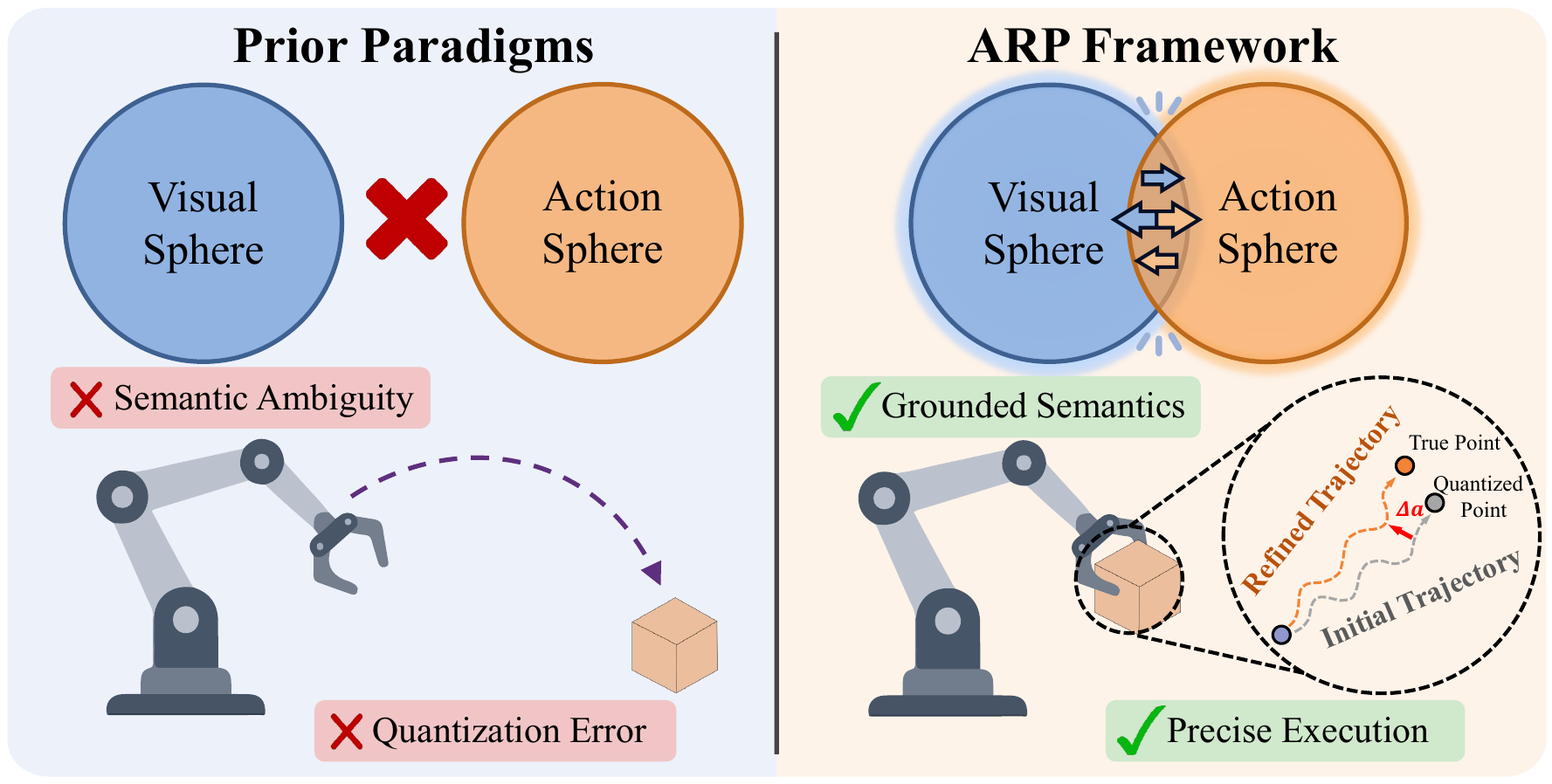}
  \vspace{-15pt}
  \caption{\textbf{ARP overview.} \textit{Left:} Prior discrete-skill policies suffer from semantic ambiguity and quantization error. \textit{Right:} ARP aligns visual embeddings with pre-quantized action latents for semantics-aware token selection and applies iterative residual refinement for precise execution.}
  \vspace{-12pt}
  \label{fig:overview}
\end{figure}

To address these challenges, we propose \textbf{Aligned Refinement Policy (ARP)}, a discrete-skill framework that couples semantic grounding with execution-level refinement for robust manipulation.
Our key insight is that action-centric tokenization offers compact and reusable motion primitives, but it introduces two bottlenecks for long-horizon multi-task control: (i) the tokens can be detached from visual semantics, causing target ambiguity under visual and language variation; and (ii) discretization discards fine-grained geometry, thus small errors accumulate and degrade precision over extended horizons.
ARP tackles these bottlenecks with two complementary mechanisms.
First, we introduce a \textbf{visual--action alignment} objective that anchors visual embeddings to pre-quantized action latents in a shared space, while keeping the skill decoder state-independent to avoid observation-conditioned overfitting.
Second, we design a lightweight \textbf{Iterative Residual Head (IRH)} that performs a minimal two-step refinement to recover fine-grained control for precise execution.
Extensive experiments show that ARP achieves state-of-the-art average results on LIBERO (89.6\%) and Meta-World (73.8\%)~\cite{libero,metaworld}, and consistently outperforms discrete-skill baselines on two long-horizon humanoid real-robot tasks on the Kuavo 4 Pro platform.

To summarize, our main contributions are threefold:
\begin{itemize}
\item \textbf{ARP framework.} A simple two-stage recipe that addresses semantic ambiguity and quantization-induced precision loss in action-centric discrete skills via alignment and lightweight refinement.
\item \textbf{Visual--action alignment.} A contrastive objective that anchors visual embeddings to action-latent representations while preserving a state-independent decoder, reducing ambiguity under visual and language variation.
\item \textbf{Iterative Residual Head (IRH).} A lightweight two-step refinement module that corrects discretization-induced errors and recovers fine-grained control.
\end{itemize}

\section{RELATED WORK}

\subsection{Generative Policies for Robot Manipulation}
Generative control policies model actions as conditional distributions, enabling expressive representations for multimodal robot behaviors. Representative directions include sequence modeling with Transformers \cite{decisiontransformer}, action tokenization with residual correction \cite{bet}, and chunked action generation for visuomotor control \cite{act}. 
Diffusion and flow-based policies broaden behavior coverage via iterative denoising, skill denoising, or flow matching under flexible conditioning \cite{diffusionpolicy,liu2024diffskill,flowpolicy}.

A practical bottleneck is iterative inference, which introduces non-trivial latency and motivates accelerated or one-step generators \cite{cp,onedp,mp1,maniflow}. Many acceleration schemes require additional consistency objectives or heavier training, increasing complexity and sensitivity. 
Instead of adopting a heavy generative refiner at inference, ARP follows the finding that combining stochasticity injection with supervised minimal iterative computation can yield strong control performance \cite{much_ado_about_noising}. In our discrete-skill setting, this motivates a two-step Iterative Residual Head (IRH) that corrects discretization-induced execution errors with low overhead.

\subsection{Discrete Skill Abstractions and Semantic Grounding}

Discrete latent skill spaces provide temporal abstraction and compositional structure for complex manipulation. Prior work explores planning in learned latent spaces \cite{tap,hgap}, variable-length skill tokenization \cite{prise}, and transformer-based policies over discrete skills \cite{vqbet,oat}. A common hierarchical paradigm learns state-independent skills in a Stage-I autoencoder and defers observation dependence to a Stage-II policy \cite{quest,star,mgp,pfdag}. While this can reduce observation-conditioned overfitting, it may detach discrete skills from visual semantics, causing semantic ambiguity.

Beyond semantic ambiguity, discrete token generation can discard fine-grained geometry; prior work therefore augments discrete plans with a one-shot continuous residual correction~\cite{bet,carp}.
Meanwhile, recent work grounds latent skills semantically, including learning semantic skill spaces via contrastive objectives~\cite{atomskill} and leveraging discrete latents for multi-task visuomotor policy learning~\cite{discrete_policy}.
Within the common two-stage skill paradigm, ARP addresses both bottlenecks: Stage-I retains a state-independent decoder and anchors visual embeddings to pre-quantized action latents via contrastive alignment, while Stage-II complements the discrete skill prior with a lightweight two-step IRH to recover execution-level precision beyond one-shot correction.

\section{METHOD}

In this section, we present the \textbf{Aligned Refinement Policy (ARP)}, a framework that enhances quantized skill abstractions through two key mechanisms: (1) contrastive visual-action alignment to ensure semantic consistency in the latent space, and (2) an Iterative Residual Head (IRH) to compensate for quantization errors via fine-grained refinement.

\begin{figure*}[t]
\centering
\includegraphics[width=0.98\textwidth]{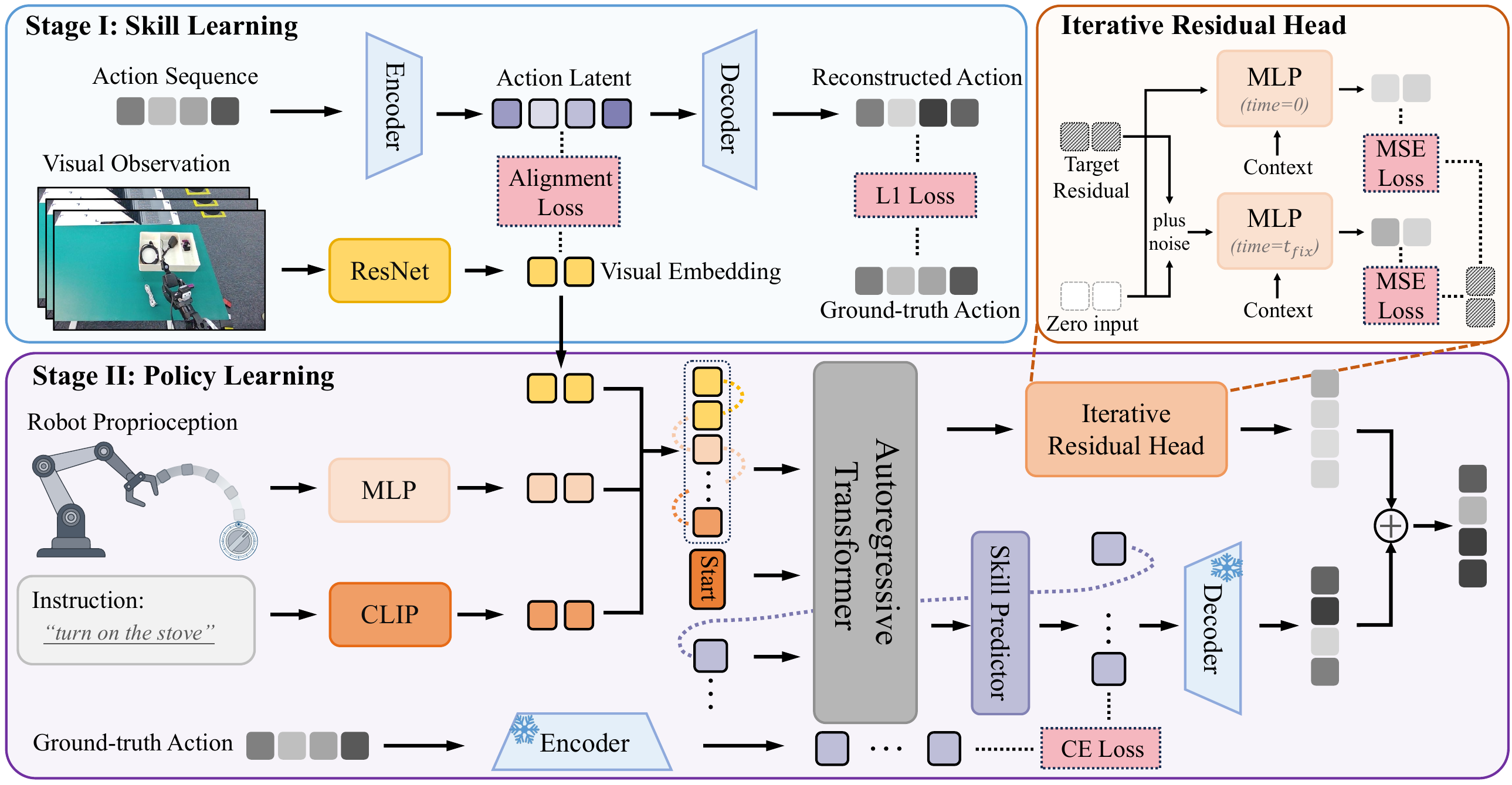}
\vspace{-3pt}
\caption{\textbf{ARP framework.} Stage I learns discrete action skills through reconstruction and contrastive alignment of action latents with visual embeddings. Stage II trains an autoregressive skill prior, decodes predicted tokens into actions, and refines them with an iterative residual head for precise execution.}
\vspace{-7pt}
\label{fig:pipeline}
\end{figure*}

\subsection{Preliminary}

\subsubsection{\textbf{Problem Formulation}} 

We consider the problem of learning a multi-task control policy from an offline dataset $\mathcal{D} = \{\tau_i\}_{i=1}^N$. 
Each trajectory $\tau$ is associated with a task description $l$ and consists of a sequence of state-action pairs $\{(s_t, a_t)\}_{t=0}^T$.
Here, $a_t \in \mathcal{A}$ denotes a continuous action vector, $s_t=(o_t,p_t)$ denotes the state, where $o_t$ is the visual observation and $p_t$ is the proprioception.
Our goal is to learn a generalizable policy $\pi(a_t | s_t, l)$ that can execute diverse tasks. 

\subsubsection{\textbf{Finite Scalar Quantization}} 
\label{sec:fsq}

To obtain discrete skill abstractions, we employ Finite Scalar Quantization (FSQ)\cite{fsq} as our discretization bottleneck. While Vector Quantization (VQ)\cite{vqvae} has been widely used in prior works, it often suffers from codebook collapse and low utilization, requiring complex heuristics (e.g., re-initialization or auxiliary losses) to stabilize training. In contrast, FSQ eliminates the need for a parameterized codebook and nearest-neighbor lookup. It projects the continuous input representation $u \in \mathbb{R}^d$ into a low-dimensional space (typically $d \in \{3, \dots, 6\}$) and quantizes it via a bounding-and-rounding operation:

\begin{equation}
z = \text{round}\left( \lfloor \mathcal{L}/2 \rfloor \odot \tanh(u) \right),
\label{eq:fsq}
\end{equation}

where $\mathcal{L} \in \mathbb{Z}^d$ specifies the number of integer quantization levels per dimension and $\odot$ denotes element-wise multiplication. 
Gradients are approximated with the Straight-Through Estimator during backpropagation. 
This mechanism removes the need for a learned codebook and nearest-neighbor assignment, and implicitly defines a discrete set of size $K=\prod_{i=1}^d \mathcal{L}_i$, avoiding codebook-collapse issues associated with VQ-style quantizers.

\subsection{Stage I: Skill Learning}

\subsubsection{\textbf{Action Tokenizer}}
Following QueST~\cite{quest}, we use a causal autoencoder to compress continuous action sequences into discrete skill tokens. 
Given $\mathbf{a} \in \mathbb{R}^{H \times d_a}$, the encoder $E_\phi$ maps it to continuous latents using causal 1D strided convolutions and masked self-attention, and each latent depends only on past actions. 
We then quantize these continuous latents into discrete skill tokens $\mathbf{z}$ via the FSQ bottleneck (Sec.~\ref{sec:fsq}), and decode $\hat{\mathbf{a}}$ with $D_\psi$:
\begin{equation}
\mathbf{z} = \text{FSQ}(E_\phi(\mathbf{a})), \quad \hat{\mathbf{a}} = D_\psi(\mathbf{z}).
\end{equation}

\subsubsection{\textbf{Visual–Action Alignment}}

A widely used design in hierarchical policy learning is to decouple skill abstraction from decision-making. Many prior works\cite{quest,carp,vqbet,discrete_policy,vqvla} deliberately restrict the first stage to learn state-independent action representations. The rationale is to treat skills as reusable motion primitives (how to move) while deferring state-dependency (when to move) to the second-stage policy. QueST \cite{quest} empirically supports this design, showing that conditioning the decoder on observations leads to performance deterioration due to overfitting to visual backgrounds.

We concur with this finding and adopt a state-independent decoder. However, we identify a critical limitation in this design: completely isolating the latent space from vision introduces semantic ambiguity. For instance, an approach/reach primitive is broadly reusable, but selecting the correct instance requires object-specific semantics, which are missing when skill latents are learned purely from actions.

To bridge this gap without sacrificing a state-independent decoder, we use contrastive learning to align visual embeddings with pre-quantized action latents. 
Specifically, we use an InfoNCE~\cite{infonce_representation} objective over matched action--vision pairs.

Formally, given a minibatch of $N$ paired samples $\{(x_i, y_i)\}_{i=1}^N$, where $x_i$ denotes the pre-quantized action latent (before FSQ) and $y_i$ is the visual embedding extracted from observations by the same vision encoder used in Stage II, we project them into a shared space via $q_i = g_q\big(\operatorname{stopgrad}(x_i)\big)$ and $k_i = g_k(y_i)$. 
The stop-gradient prevents the alignment objective from updating the action quantizer, preserving reusable motion primitives while anchoring visual embeddings to action latents for robust Stage-II routing. 
The loss for the $i$-th pair is:

\begin{equation}
\mathcal{L}_{\text{NCE}, i} = - \log \frac{\exp(\text{sim}(q_i, k_i) / \tau)}{\sum_{j=1}^N \exp(\text{sim}(q_i, k_j) / \tau)},
\label{eq:infoNCE}
\end{equation}
where the denominator includes one positive pair and $N-1$ negatives (in-batch negatives), $\text{sim}(\cdot)$ denotes cosine similarity, and $\tau$ is a temperature hyperparameter. 
In practice, we use a symmetric variant by computing InfoNCE in both directions and averaging:

\begin{equation}
\mathcal{L}_{align}=\frac{1}{2}\left(\mathcal{L}_{q\rightarrow k}+\mathcal{L}_{k\rightarrow q}\right),
\label{eq:l_align}
\end{equation}

where $\mathcal{L}_{q\rightarrow k}$ uses $q$ as queries and $k$ as keys, and $\mathcal{L}_{k\rightarrow q}$ swaps them.
This aligns visual representations with action-latent representations in a shared embedding space, improving semantics-aware token selection without injecting observations into the decoder.

\subsubsection{\textbf{Training Objective}}  
We combine the reconstruction loss with alignment regularization. The total loss is:

\begin{equation}
\mathcal{L}_{Stage1} = \mathcal{L}_{recon} + \lambda \mathcal{L}_{align},
\end{equation}
where $\mathcal{L}_{recon} = \|\mathbf{a} - \hat{\mathbf{a}}\|_1$ is the L1 reconstruction loss following \cite{quest}, and $\mathcal{L}_{align}$ is the InfoNCE loss in Eq.~(\ref{eq:l_align}).

\subsection{Stage II: Policy Learning}

\subsubsection{\textbf{Autoregressive Skill Prior}} 

After training the Stage 1 autoencoder, we freeze its weights and train a latent skill prior $\pi_\theta(\mathbf{z} | s_{t-h:t}, e)$ to model the distribution of skill tokens conditioned on the state history and task specification, where $e$ is the embedding of the task description $l$. 
To effectively aggregate multi-modal context, we encode the visual history $o_{t-h:t}$ and proprioceptive states $p_{t-h:t}$ via modality-specific encoders and concatenate the resulting tokens with a task embedding $e$ to form the context sequence $C_t$.
Conditioned on $C_t$, a decoder-only Transformer then predicts the skill sequence $\mathbf{z} = (z^1, \dots, z^L)$ autoregressively, attending to both the causal token history and the context $C_t$:

\begin{equation}
\pi_\theta(\mathbf{z} | C_t) = \prod_{i=1}^L \pi_\theta(z^i | \texttt{<s>}, z^{<i}, C_t),
\end{equation}

where $\texttt{<s>}$ is a start token. The prior is optimized via standard cross-entropy loss against the ground-truth tokens $\mathbf{z}_{gt}$ extracted by the frozen Stage 1 encoder:

\begin{equation}
\mathcal{L}_{prior} = - \mathbb{E}_{\tau \sim \mathcal{D}}
\left[\sum_{i=1}^L \log \pi_\theta\!\left(z^i_{\mathrm{gt}} \mid z^{<i}_{\mathrm{gt}}, C_t\right)\right].
\end{equation}
\subsubsection{\textbf{Iterative Residual Head}}

\begin{algorithm}[tbp]
\caption{Two-step IRH: training and inference}
\label{alg:irh}
\begin{algorithmic}[1]
\REQUIRE Context features $C_t$, target residual $\Delta a$, fixed refinement step $\rho_{\mathrm{fix}}$, noise std $\sigma$, weights $w_0,w_1$
\STATE \textbf{Training:}
\STATE \hspace{0.6em}$x_0 \leftarrow \mathbf{0}$
\STATE \hspace{0.6em}Sample $\epsilon \sim \mathcal{N}(0,I)$
\STATE \hspace{0.6em}$x_{\rho_{\mathrm{fix}}} \leftarrow \Delta a + (1-\rho_{\mathrm{fix}})\sigma\epsilon$
\STATE \hspace{0.6em}$\widehat{\Delta a}_{0} \leftarrow R_\omega(x_0,0,C_t)$
\STATE \hspace{0.6em}$\widehat{\Delta a}_{1} \leftarrow R_\omega(x_{\rho_{\mathrm{fix}}},\rho_{\mathrm{fix}},C_t)$
\STATE \hspace{0.6em}$\mathcal{L}_{refine} \leftarrow w_0\|\widehat{\Delta a}_{0}-\Delta a\|_2^2 + w_1\|\widehat{\Delta a}_{1}-\Delta a\|_2^2$
\STATE \hspace{0.6em}\textbf{return} $\mathcal{L}_{refine}$
\STATE \textbf{Inference:}
\STATE \hspace{0.6em}$\widehat{\Delta a}_{0} \leftarrow R_\omega(\mathbf{0},0,C_t)$
\STATE \hspace{0.6em}$\widehat{\Delta a}_{1} \leftarrow R_\omega(\widehat{\Delta a}_{0},\rho_{\mathrm{fix}},C_t)$
\STATE \hspace{0.6em}\textbf{return} $\widehat{\Delta a}_{1}$
\end{algorithmic}
\end{algorithm}

While discrete tokens capture high-level structure, quantization errors limit precision. To address this, we introduce the Iterative Residual Head (IRH).

We explored alternatives including one-shot MLP regression, diffusion~\cite{ddpm}, and flow matching~\cite{flow_matching}. 
Inspired by \cite{much_ado_about_noising}, which shows that a minimal number of supervised refinement steps, when paired with training-time stochasticity, can match much of the benefit of generative policies, we adopt a stabilized two-step refinement scheme.

Let $a_{\mathrm{base}}=D_\psi(\mathbf{z})$ be the action decoded from predicted tokens, and define the residual target as $\Delta a = a_{\mathrm{gt}}-a_{\mathrm{base}}$.
IRH predicts this residual with a conditional network $R_\omega(x,\rho,C_t)$, where $x$ is the residual seed and $\rho$ denotes the refinement step.
The training objective and inference procedure are summarized in Alg.~\ref{alg:irh}.

During training, we minimize $\mathcal{L}_{refine}=w_0\|\widehat{\Delta a}_{0}-\Delta a\|_2^2+w_1\|\widehat{\Delta a}_{1}-\Delta a\|_2^2$, and at inference, IRH outputs $a_{\mathrm{final}}=a_{\mathrm{base}}+\widehat{\Delta a}_1$ after two refinement steps.

\subsubsection{\textbf{Training Objective}}
We jointly train the prior and the residual head. The Stage-II objective is
\begin{equation}
\mathcal{L}_{Stage2} = \mathcal{L}_{prior} + \lambda(n)\,\mathcal{L}_{refine}.
\end{equation}
Since the residual head relies on base actions from the prior, we warm up the refinement weight with a logistic schedule
$\lambda(n)=\left(1+e^{-k(n-t_0)}\right)^{-1}$,
where $n$ is the training epoch and $k,t_0$ control the slope and midpoint.
This curriculum downweights refinement early, allowing the prior to learn reliable base actions before emphasizing residual corrections.

\section{SIMULATION EXPERIMENTS}

\begin{figure}[t]
\centering
\includegraphics[width=0.98\linewidth]{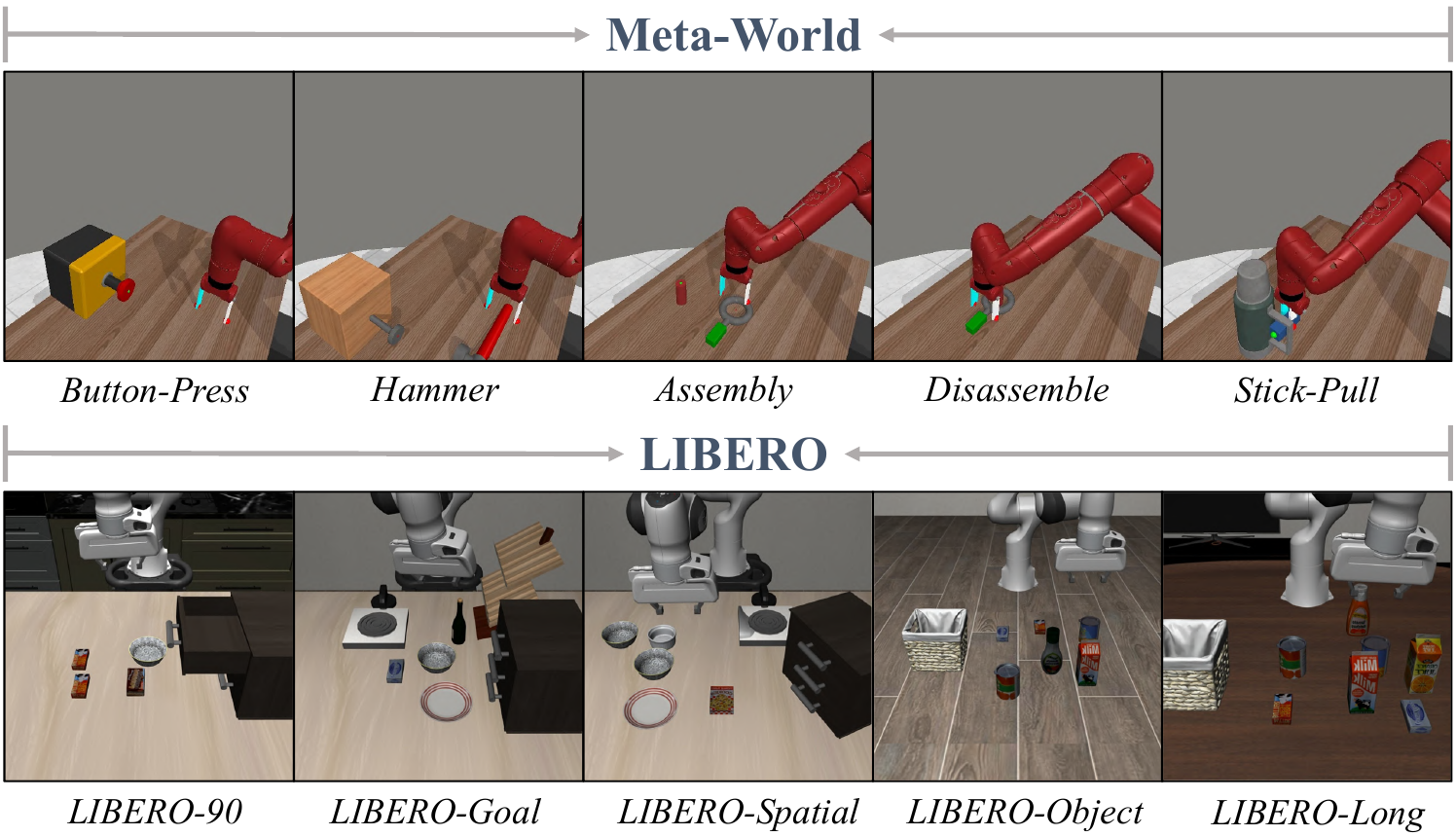}
\vspace{-5pt}
\caption{Sample visualizations of tasks from Meta-World and LIBERO.}
\vspace{-8pt}
\label{fig:sim_bar}
\end{figure}

\subsection{Benchmarks and Implementation Details}

\subsubsection{\textbf{Simulation Benchmarks}}

We evaluate ARP on two widely used robotic manipulation benchmarks:
\textbf{Meta-World~\cite{metaworld}:} A suite of 50 manipulation tasks for fine-grained continuous control. We follow the standard single-task setting and generate ten expert demonstrations per task using Meta-World scripted expert policies.
\textbf{LIBERO~\cite{libero}:} A large-scale benchmark with 130 tasks organized into five suites: LIBERO-90, Spatial, Object, Goal, and Long, designed for multi-task generalization and long-horizon reasoning. For each suite, we use the officially provided 50 demonstrations per task and train a single multi-task policy. 
Sample visualizations of tasks are shown in Fig.~\ref{fig:sim_bar}.

\begin{table*}[htbp] 
\centering
\caption{
LIBERO Overall Performance.
Baselines marked with $^\dag$ are cited from original papers.
Success Rate (\%).
}
\vspace{-5pt}
\small
\resizebox{0.98\linewidth}{!}{
\begin{tabular}{lccccc|c}
\toprule[1.25pt]
\textbf{Method} & \textbf{LIBERO-Object} & \textbf{LIBERO-Spatial} & \textbf{LIBERO-Goal} & \textbf{LIBERO-Long} & \textbf{LIBERO-90} & \textbf{Overall} \\
\midrule
Octo$^\dag$\cite{octo}       & \score{85.7}{0.9} & \score{78.9}{1.0} & \score{84.6}{0.9} & \score{51.1}{1.3} & --                & \score{75.1}{0.6} \\
OpenVLA$^\dag$\cite{openvla}    & \score{88.4}{0.8} & \score{84.7}{0.9} & \score{79.2}{1.0} & \score{53.7}{1.3} & --                & \score{76.5}{0.6} \\
TraceVLA$^\dag$\cite{tracevla}     & \score{85.2}{0.4} & \score{84.6}{0.2} & \score{75.1}{0.3} & \score{74.8}{1.0} & --                & \score{74.8}{0.5} \\
SpatialVLA$^\dag$\cite{spatialvla}     & \score{89.9}{0.7} & \score{88.2}{0.5} & \score{78.6}{0.6} & \score{55.5}{1.0} & --                & \score{78.1}{0.7} \\
\midrule
ResNet-T\cite{libero}           & \score{43.4}{0.8} & \score{82.1}{0.7} & \score{83.8}{1.6} & \score{49.0}{2.9} & \score{83.7}{0.9} & \score{68.4}{0.6} \\
Diffusion Policy\cite{diffusionpolicy}   & \score{78.2}{0.6} & \score{71.9}{0.9} & \score{73.6}{2.8} & \score{45.8}{4.3} & \score{74.6}{0.8} & \score{68.8}{0.4} \\
ACT\cite{act}                & \score{58.6}{0.6} & \score{77.9}{2.7} & \score{70.3}{3.5} & \score{48.0}{2.6} & \score{55.3}{1.1} & \score{62.0}{1.7} \\
\midrule
MGP$^\dag$\cite{mgp} & -- & -- & -- & \scoremean{77.0} & \scoremean{88.9} & -- \\
VQ-BeT\cite{vqbet}             & \score{77.1}{2.7} & \score{84.0}{0.8} & \score{64.8}{0.4} & \score{59.7}{0.4} & \score{81.8}{0.5} & \score{73.5}{0.5} \\
QueST\cite{quest}              & \score{80.7}{1.5} & \score{79.5}{0.7} & \score{81.0}{1.0} & \score{70.0}{0.8} & \score{86.2}{0.5} & \score{79.5}{0.4} \\

\rowcolor[rgb]{.902,.996,1.0}
\textbf{ARP(Ours)}               & \bestscore{92.8}{1.2} & \bestscore{91.6}{0.2} & \bestscore{88.8}{0.2} & \bestscore{83.6}{0.2} & \bestscore{91.0}{0.6} & \bestscore{89.6}{0.2} \\
\bottomrule[1.1pt]
\end{tabular}
}
\label{tab:libero_overall_performance_1}
\vspace{-10pt}
\end{table*}

\subsubsection{\textbf{Baselines}}
We benchmark ARP against a diverse set of strong and representative continuous-action and discrete-skill policies. \textbf{Diffusion \& Continuous Methods.} We include diffusion models such as Diffusion Policy (DP) \cite{diffusionpolicy} and its 3D variant DP3 \cite{dp3}, accelerated frameworks including Consistency Policy (CP) \cite{cp}, FlowPolicy \cite{flowpolicy}, and Action Chunking with Transformers (ACT) \cite{act}. \textbf{Discrete \& Autoregressive Methods.} We compare token-based and autoregressive architectures, including Vector-Quantized Behavior Transformer (VQ-BeT) \cite{vqbet}, Quantized Skill Transformer (QueST) \cite{quest}, and Masked Generative Policy (MGP) \cite{mgp}. 
Where available, we additionally report Vision-Language-Action (VLA) model results on LIBERO from prior work.

\subsubsection{\textbf{Evaluation Metrics}}
For fair evaluation, all experiments are conducted with three random seeds, and we report the mean success rate (SR) across seeds.
For Meta-World, following~\cite{mgp}, we evaluate the policy over 20 episodes every 100 training iterations and compute the average of the top-5 success rates per task across training. For LIBERO, SR is measured over 50 evaluation episodes per suite.

\subsection{Performance Comparison}

\begin{table}[t]
\setlength{\tabcolsep}{2pt}
\renewcommand{\arraystretch}{0.9}
\centering
\caption{Performance on Meta-World. Success Rate (\%).}
\vspace{-5pt}
\label{tab:Meta-World}
\begin{tabular}{lccccc}
\toprule
\multirow{2}{*}{Methods} & \multicolumn{5}{c}{Meta-World} \\
\cmidrule(lr){2-6}
 & Easy (28) & Medium (11) & Hard (6) & Very Hard (5) & Avg. SR \\
\midrule
DP\cite{diffusionpolicy}        & 83.6 & 31.1 & 10.8 & 26.6 & 38.0 \\
DP3\cite{dp3}                  & 90.9 & 61.6 & 38.0 & 49.0 & 59.9 \\
CP\cite{cp}                    & 91.2 & 62.7 & 40.0 & 51.0 & 61.2 \\
FlowPolicy\cite{flowpolicy}    & 90.2 & 63.0 & 39.2 & 36.0 & 57.1 \\
MGP\cite{mgp}                  & \textbf{92.0} & 65.0 & 44.0 & 53.8 & 63.7 \\
\rowcolor[rgb]{.902,.996,1.0}
\textbf{ARP(Ours)}                  & 91.5 & \textbf{71.3} & \textbf{65.8} & \textbf{67.9} & \textbf{73.8} \\
\bottomrule
\end{tabular}
\vspace{-8pt}
\end{table}

\subsubsection{\textbf{Multi-Task Evaluation on LIBERO}}

Table~\ref{tab:libero_overall_performance_1} reports results on all 130 tasks across five LIBERO suites. 
ARP achieves the best overall success rate (89.6\%), outperforming both continuous-action and discrete-skill baselines.
The gains are especially clear on LIBERO-Object and LIBERO-Long, where language-conditioned target selection and long-horizon composition are critical: ARP reaches 92.8\% on Object and 83.6\% on Long, improving over QueST by +12.1\% and +13.6\%, respectively.
These results suggest that visual--action alignment improves semantic skill selection, while IRH supports robust execution after the correct skill is selected.

\subsubsection{\textbf{Single-Task Evaluation on Meta-World}}

Table~\ref{tab:Meta-World} reports single-task results on 50 Meta-World tasks. 
ARP achieves the best average success rate (73.8\%), exceeding the strongest baseline MGP by +10.1\%.
The improvement is concentrated on precision-critical tasks: ARP reaches 65.8\% on Hard and 67.9\% on Very Hard, outperforming MGP by +21.8\% and +14.1\%, respectively.
This trend is consistent with IRH reducing discretization-induced errors and recovering fine-grained control.

\begin{table}[t]
\centering
\caption{Ablation study on core components.}
\vspace{-5pt}
\label{tab:causal_ablation}
\begin{tabular}{lcc c}
\toprule
\multirow{2}{*}{Task} & \multicolumn{2}{c}{Core Components} & \multirow{2}{*}{Success Rate (\%)} \\
\cmidrule(lr){2-3}
 & V--A Align. & IRH &  \\
\midrule
\multirow{4}{*}{LIBERO-Long}
 & $\times$ & $\times$ & \score{71.1}{1.5} \\
 & $\checkmark$ & $\times$ & \score{74.1}{0.5} \\
 & $\times$ & $\checkmark$ & \score{75.5}{0.3} \\
 & $\checkmark$ & $\checkmark$ & \bestscore{83.6}{0.2} \\
\midrule
\multirow{4}{*}{Meta-World(Hard)}
 & $\times$ & $\times$ & \score{40.7}{0.8} \\
 & $\checkmark$ & $\times$ & \score{56.9}{0.4} \\
 & $\times$ & $\checkmark$ & \score{51.2}{0.5} \\
 & $\checkmark$ & $\checkmark$ & \bestscore{65.8}{1.6} \\
\bottomrule
\end{tabular}
\vspace{-5pt}
\end{table}

\subsection{Ablation Studies}

To systematically evaluate the design choices in our ARP framework, we conduct extensive ablation studies on two representative environments: LIBERO-Long (testing long-horizon semantic reasoning) and Meta-World(Hard) (testing precision-critical manipulation).

\subsubsection{\textbf{Effectiveness of Core Components}}
To evaluate each key component in ARP, we ablate the two core modules---Stage-I visual--action alignment and Stage-II residual refinement (IRH). Results are summarized in Table~\ref{tab:causal_ablation}.

On LIBERO-Long, enabling alignment alone improves the success rate from 71.1\% to 74.1\% (+3.0\%), suggesting that alignment anchors the visual embedding space to the action-latent space and mitigates semantic ambiguity under visual/language variations. Enabling residual refinement alone yields 75.5\% (+4.4\%), indicating that even when the discrete plan is reasonable, quantization can still limit overall execution fidelity. Combining both achieves 83.6\% (+12.5\%), demonstrating strong complementarity between correct skill selection and precise execution.

On Meta-World(Hard), alignment yields a substantial gain from 40.7\% to 56.9\% (+16.2\%), emphasizing observation grounding in precision-critical manipulation. Residual refinement boosts performance to 51.2\% (+10.5\%), and the full model reaches 65.8\% (+25.1\%). This suggests challenging tasks benefit from coupling semantically grounded discrete skills with continuous residual corrections to bridge the gap between high-level planning and precise control.

\begin{table}[t]
\centering
\caption{Comparison of skill abstraction designs. Success Rate (\%).}
\vspace{-5pt}
\label{tab:quant_compare}
\footnotesize
\setlength{\tabcolsep}{4pt}
\begin{tabular}{lcccc}
\toprule
 & VQ & LFQ  & Obs-Cond. & Ours (FSQ) \\
\midrule
LIBERO-Long     & \score{74.9}{2.7} & \score{73.3}{0.9} & \score{68.7}{0.5} & \bestscore{83.6}{0.2} \\
Meta-World(Hard)  & \score{62.1}{1.2} & \score{63.2}{0.5} & \score{62.4}{0.3} & \bestscore{65.8}{1.6} \\
\bottomrule
\end{tabular}
\end{table}

\begin{figure}[t]
\centering
\includegraphics[width=0.98\linewidth]{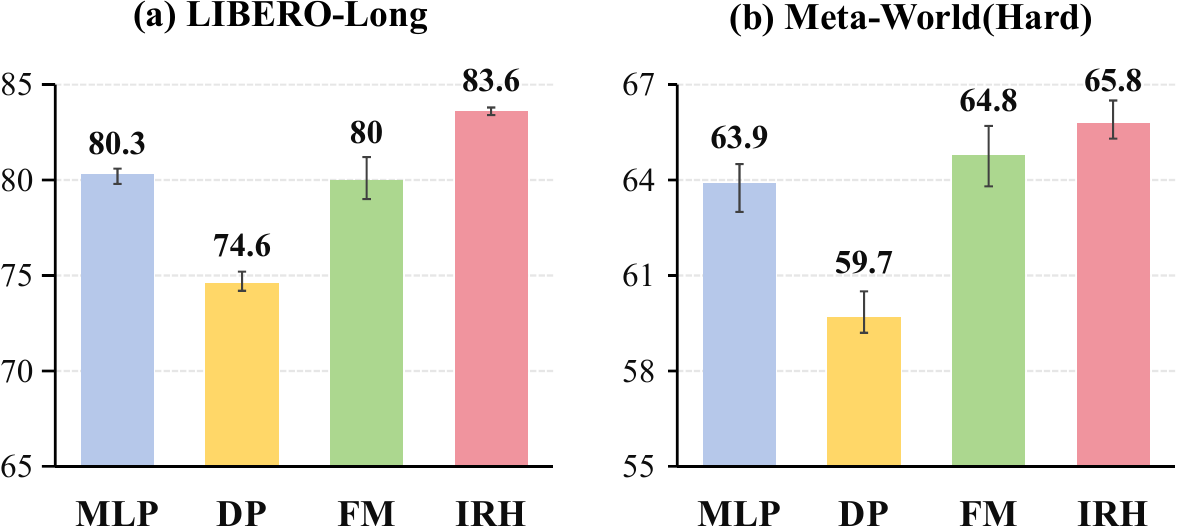}
\vspace{-3pt}
\caption{Comparison of refinement strategies and residual head design.}
\vspace{-10pt}
\label{fig:residual_head}
\end{figure}

\subsubsection{\textbf{Design Choices in Skill Abstraction}}

Table~\ref{tab:quant_compare} analyzes two design axes in skill abstraction:
(i) Quantization: we compare Vector Quantization (VQ)\cite{vqvae}, Lookup-Free Quantization (LFQ)\cite{lfq}, and Finite Scalar Quantization (FSQ)\cite{fsq} under the same contrastive alignment objective,  with all else fixed.
(ii) Adding visual semantics: using the same FSQ bottleneck, we compare latent-level alignment to an \textit{Obs-Cond.} baseline that conditions the Stage-I action decoder on observations instead of using contrastive alignment.

\noindent \textbf{Comparison of Quantization Strategies.} Replacing FSQ with VQ or LFQ degrades performance, especially on LIBERO-Long (83.6\% vs.\ 74.9\%/73.3\%). One possible reason is that FSQ's multi-level scalar discretization yields more stable and fine-grained action tokens~\cite{fsq}, whereas VQ can suffer from codebook under-utilization and LFQ relies on binary independent latents with additional regularization~\cite{lfq}, which may be less robust under long-horizon rollouts.

\noindent\textbf{Alignment vs. observation-conditioned decoding.} Conditioning the Stage-I action decoder on observations (\textit{Obs-Cond.}) reduces LIBERO-Long performance to 68.7\%, consistent with observation-conditioned decoding overfitting to spurious visual backgrounds. In contrast, latent alignment anchors visual embeddings to action latents while preserving a state-independent decoder, improving generalization.

\begin{table}[t]
\centering
\caption{IRH iterations ablation. Success Rate (\%).}
\vspace{-5pt}
\label{tab:irh_steps}
\setlength{\tabcolsep}{3pt}
\footnotesize
\begin{tabular}{lcccc}
\toprule
 & 1 Step(MLP) & 2 Steps(IRH) & 3 Steps & 5 Steps \\
\midrule
LIBERO-Long & \score{80.3}{0.4} & \score{83.6}{0.2} & \score{83.8}{0.6} & \score{83.5}{0.8} \\
\bottomrule

\end{tabular}
\vspace{-10pt}
\end{table}

\subsubsection{\textbf{Comparison of Residual Refinement Strategies}}

To justify our Stage-II refinement design, we compare our Iterative Residual Head (IRH) with three alternatives: a deterministic MLP (MLP), a diffusion-based head (DP), and a flow-matching head (FM). Results are shown in Fig.~\ref{fig:residual_head}.

\noindent \textbf{When generative refinement is inefficient.} The diffusion-based head is the weakest on both benchmarks (74.6\% on LIBERO-Long and 59.7\% on Meta-World(Hard)). Diffusion models are powerful for modeling full action distributions, but applying multi-step denoising to predict a small residual $\Delta a$ is a poor computational match and adds sensitivity to the denoising schedule. The FM head is substantially stronger (80.0\% and 64.8\%), but still slightly behind IRH.

\noindent \textbf{Why minimal iterative refinement works best.} Compared to one-step refinement (MLP), IRH performs a lightweight two-step correction inspired by~\cite{much_ado_about_noising}. IRH improves performance to 83.6\% on LIBERO-Long and 65.8\% on Meta-World(Hard), exceeding the one-step MLP by +3.3\% and +1.9\%, respectively, suggesting that a second noise-aware refinement step provides additional correction capacity for contact-sensitive precision.
We further test deeper refinement and observe negligible gains ($<0.3\%$) (Table~\ref{tab:irh_steps}), indicating that performance saturates quickly and two iterations suffice.
Overall, IRH achieves the best success rate with only two refinement steps, whereas diffusion/flow-based heads require substantially more steps.

\begin{figure}[t]
\centering
\includegraphics[width=0.95\linewidth]{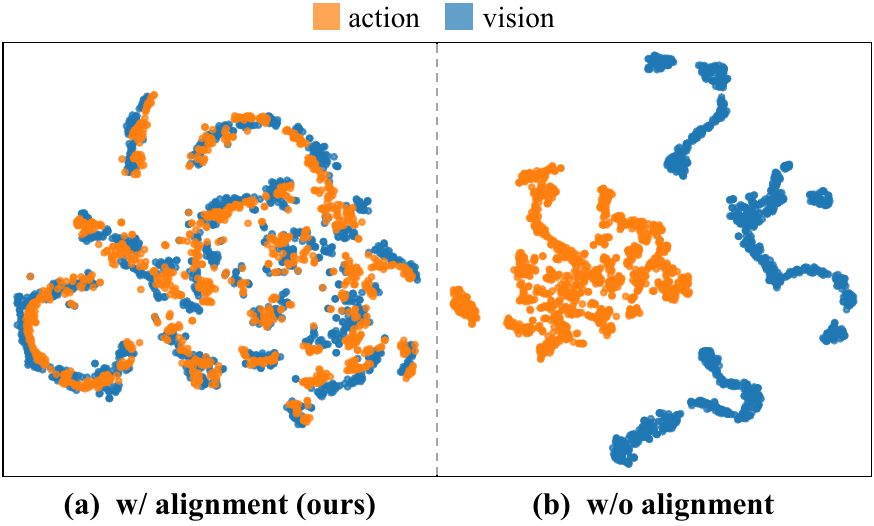}
\vspace{-3pt}
\caption{t-SNE visualization of the visual embeddings (blue) and pre-quantized action latents (orange).}
\vspace{-10pt}
\label{fig:tsne}
\end{figure}

\begin{figure*}[t!]
\centering
\includegraphics[width=0.98\textwidth]{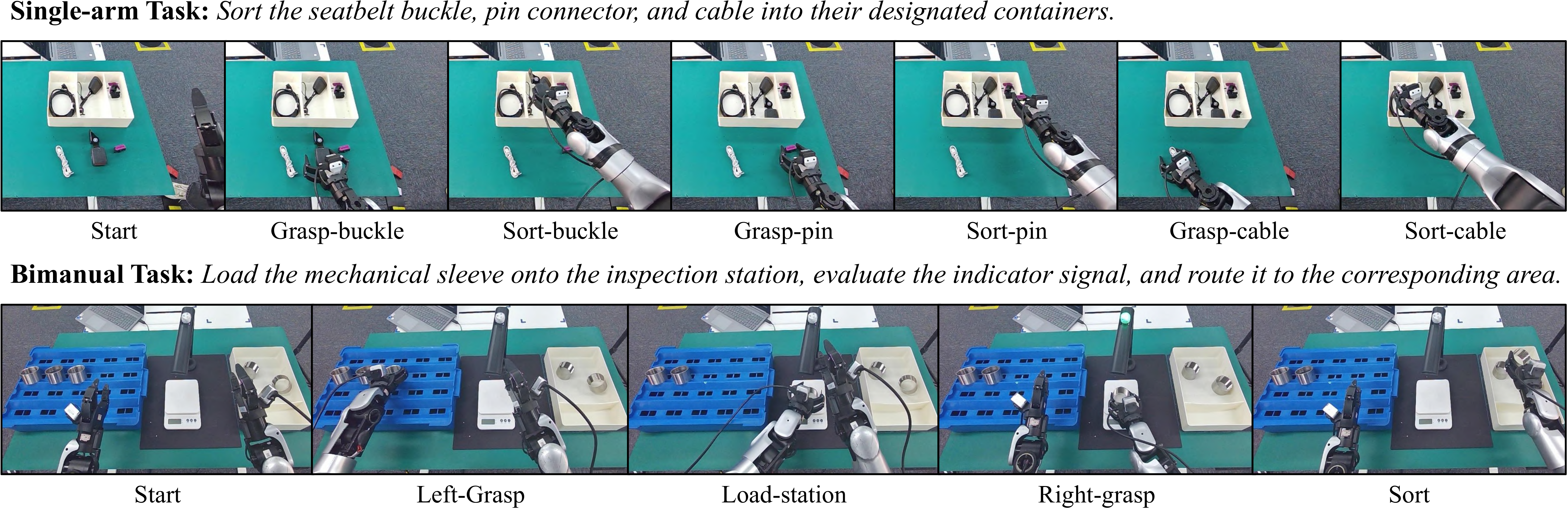}
\vspace{-5pt}
\caption{Visualization of two real-world manipulation tasks with key stages. Each frame highlights a critical step in the manipulation sequence.}
\vspace{-12pt}
\label{fig:real_world}
\end{figure*}
\subsubsection{\textbf{Latent Space Representation Analysis}}

To visualize the effect of Stage-I contrastive alignment, we apply t-SNE to paired pre-quantized action latents and their corresponding visual embeddings in Fig.~\ref{fig:tsne}, comparing \textit{without alignment} and \textit{with alignment} under the same settings.

Without alignment, action latents (orange) and visual embeddings (blue) form disjoint clusters, suggesting a mismatch between the visual and action-latent spaces. With alignment, action/vision embeddings become co-located, suggesting that InfoNCE learns a shared space that anchors visual representations to action latents while keeping the decoder state-independent, improving semantics-aware token selection.

\section{REAL-WORLD EXPERIMENTS}

\subsection{Real Robot Benchmark}
\label{sec:real_robot}

We conduct physical experiments on the \textbf{Kuavo 4 Pro humanoid robot} platform. The real-world setting poses additional challenges, including noisy visual observations, modeling errors in physical dynamics, and complex multi-arm coordination.
For each task, we collect expert demonstrations via teleoperation to provide high-quality supervision. To evaluate generalization, the initial positions of the manipulated objects are randomized during inference. Each task is evaluated over 10 trials, and we report success counts (out of 10) as the primary metric. We design two long-horizon tasks to evaluate semantic understanding, sequential execution, and conditional reasoning. The physical environment and key execution stages for both tasks are visualized in Fig.~\ref{fig:real_world}.

\begin{itemize}
\item \textbf{Task 1 Sequential Material Sorting.}
Pick and sort three components (seatbelt buckle, cable, pin connector) into their corresponding boxes.

\item \textbf{Task 2 Bimanual Quality Inspection.}
Left arm places a mechanical sleeve on an inspection station; based on the indicator light (pass/fail), right arm transfers the sleeve to the accept/reject area.
\end{itemize}

\begin{table}[t]
\centering
\caption{Real-world results. Success counts over 10 trials. Total denotes full-task success.}
\vspace{-5pt}
\footnotesize
\setlength{\tabcolsep}{2pt}
\renewcommand{\arraystretch}{1.1}
\begin{tabular}{lcccc@{\hspace{4pt}}ccc}
\toprule
\multirow{2}{*}{Method} 
& \multicolumn{4}{c}{Task 1} 
& \multicolumn{3}{c}{Task 2} \\
\cmidrule(lr){2-5}\cmidrule(lr){6-8}
& Buckle & Cable & \makecell{Pin\\Connector} & Total 
& \makecell{Sleeve$\rightarrow$\\Station} & \makecell{Station$\rightarrow$\\Bin} & Total \\
\midrule
VQ-BeT      & 8/10 & 7/10 & 5/10 & 5/10 & 1/10 & 4/10 & 1/10 \\
QueST       & 5/10 & 7/10 & 4/10 & 3/10 & 2/10 & 5/10 & 2/10 \\
\rowcolor[rgb]{.902,.996,1.0}
\textbf{ARP(Ours)}  & \textbf{9/10} & \textbf{8/10} & \textbf{7/10} & \textbf{7/10} & \textbf{4/10} & \textbf{8/10} & \textbf{4/10} \\
\bottomrule
\end{tabular}
\label{tab:real_results}
\vspace{-10pt}
\end{table}

\subsection{Performance Comparison}

We benchmark ARP against two strong discrete-skill baselines, VQ-BeT~\cite{vqbet} and QueST~\cite{quest} (Table~\ref{tab:real_results}).

\subsubsection{\textbf{Task 1 Sequential Material Sorting}}
ARP achieves the highest end-to-end success rate of 7/10, outperforming VQ-BeT (5/10) and QueST (3/10), while maintaining strong per-object success (9/10, 8/10, and 7/10).
The gains are consistent with residual refinement mitigating discretization-induced errors. 
Stage-I alignment stabilizes perception-conditioned skill selection under visual variation, while IRH provides corrections to recover execution-level precision.

\subsubsection{\textbf{Task 2 Bimanual Quality Inspection}}
Task 2 is more challenging due to bimanual coordination and perception-conditioned routing, leading to lower end-to-end success rates for all methods. ARP achieves 4/10 overall success, compared to QueST (2/10) and VQ-BeT (1/10).
The stage-wise breakdown provides further insight. In the \textit{Sleeve$\rightarrow$Station} phase, ARP improves success (4/10 vs.\ 2/10 and 1/10), suggesting that residual refinement helps with placement under unmodeled dynamics. In the \textit{Station$\rightarrow$Bin} phase, ARP achieves 8/10, indicating reliable indicator-conditioned routing, consistent with the benefit of a semantics-aware routing interface for perception-driven branching decisions.

\subsection{Failure Case Analysis}

\begin{figure}
    \centering
    \includegraphics[width=0.98\linewidth]{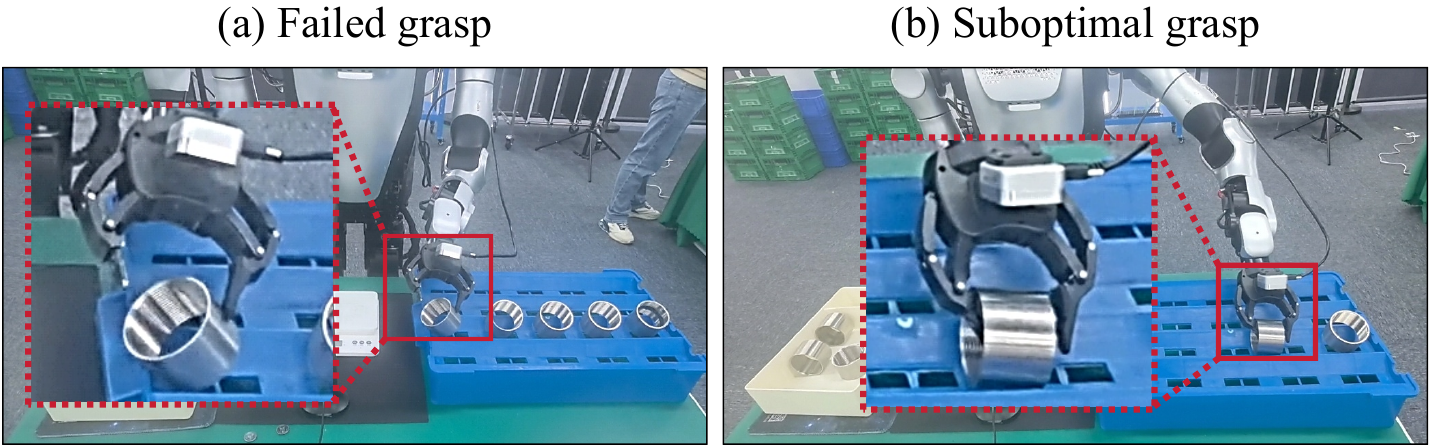}
    \vspace{-3pt}
    \caption{Failure cases on Kuavo~4~Pro}
    \vspace{-12pt}
    \label{fig:fail_case}
\end{figure}

We observe two recurring failure modes in contact-sensitive manipulation on Kuavo~4~Pro (Fig.~\ref{fig:fail_case}).
\textbf{(a) Failed grasp.} Small pose/contact or friction variations can prevent stable contact, and the policy often lacks recovery, leading to slippage.
\textbf{(b) Suboptimal grasp.} The object is lifted, but an incorrect grasp pose degrades downstream execution and may cause task failure.
Overall, these cases suggest ARP may need stronger closed-loop recovery and finer, millimeter-level contact-aware control for diverse objects.

\section{Conclusion}

In this work, we present Aligned Refinement Policy (ARP), a two-stage discrete-skill framework that enhances quantized skills with both semantic grounding and execution-level precision.
Stage I contrastively aligns visual embeddings with pre-quantized action latents while keeping the decoder state-independent.
Stage II adds a lightweight two-step Iterative Residual Head (IRH) to correct quantization errors.
Across Meta-World, LIBERO, and a Kuavo 4 Pro humanoid robot, ARP outperforms strong discrete-skill baselines and improves long-horizon manipulation in both simulation and the real world.
We believe visual--action alignment plus minimal iterative refinement is a simple, broadly applicable recipe for robust multi-task manipulation, and future work will extend ARP to broader robots and real-world settings.

\bibliographystyle{IEEEtran}
\bibliography{references}

\end{document}